\pdfoutput=1
\documentclass[11pt]{article}

\usepackage[]{acl}

\usepackage{times}
\usepackage{latexsym}

\usepackage[T1]{fontenc}
\usepackage[utf8]{inputenc}

\usepackage{microtype}

\usepackage{inconsolata}
\usepackage{graphicx}
\usepackage{amsmath}
\usepackage{comment}
\usepackage{amssymb}
\usepackage{booktabs}
\usepackage{enumitem}
\title{Improving Socratic Question Generation using Data Augmentation and Preference Optimization}

\author{Nischal Ashok Kumar \and Andrew Lan \\
  University of Massachusetts Amherst \\
  \texttt{\{nashokkumar, andrewlan\}@cs.umass.edu}
  }

\begin{document}
\maketitle
\begin{abstract}
The Socratic method is a way of guiding students toward solving a problem independently without directly revealing the solution to the problem by asking incremental questions. Although this method has been shown to significantly improve student learning outcomes, it remains a complex labor-intensive task for instructors. Large language models (LLMs) can be used to augment human effort by automatically generating Socratic questions for students. However, existing methods that involve prompting these LLMs sometimes produce invalid outputs, e.g., those that directly reveal the solution to the problem or provide irrelevant or premature questions. To alleviate this problem, inspired by reinforcement learning with AI feedback (RLAIF), we first propose a data augmentation method to enrich existing Socratic questioning datasets with questions that are invalid in specific ways. Also, we propose a method to optimize open-source LLMs such as LLama 2 to prefer ground-truth questions over generated invalid ones, using direct preference optimization (DPO). Our experiments on a Socratic questions dataset for student code debugging show that a DPO-optimized LLama 2-7B model can effectively avoid generating invalid questions, and as a result, outperforms existing state-of-the-art prompting methods\footnote{The code for our paper can be found at: \url{https://github.com/umass-ml4ed/socratic-quest-gen}}. 
\end{abstract}

\section{Introduction}

Learning based on a conversation that consists of questions and answers, where the student responds to questions posed by a more knowledgeable instructor, has been proven to be effective in teaching students about a particular concept \cite{wood1976role}. In particular, \emph{Socratic questioning}, which refers to a way for the instructor to guide a student to solve a problem (within their zone of proximal development) by asking them questions that promote thinking while not directly revealing the solution \cite{quintana2018scaffolding}, is a very effective pedagogical method in conversation-based learning and tutoring. 

Recent advances in large language models (LLMs) \cite{bubeck2023sparks} have led to the rapid development of chatbots that promote student learning by automatically generating the instructor's utterances \cite{dan2023educhat,kazemitabaar2024codeaid,tanwar2024opinebot}. One key area of interest in the development of such chatbots is question generation, which can help students solve logical problems in the mathematics and programming domains \cite{al-hossami-etal-2023-Socratic,shridhar-etal-2022-automatic}. Typically, question generation in educational applications has focused on generating practice or assessment questions, in biology exams \cite{wang2018qg}, reading comprehension \cite{ashok-kumar-etal-2023-improving}, math practice \cite{wang-etal-2021-math}, and programming exercises \cite{sarsa2022automatic}. As a specific form of question generation, Socratic question generation has gained attention, owing to its effectiveness in improving student learning outcomes by eliciting critical thinking and self-discovery during problem-solving \cite{paul2007critical}. 

Socratic questions generation is a complex task because it involves %questions based on a fixed input \cite{wang2018qg,ashok-kumar-etal-2023-improving,sarsa2022automatic} as the 
mapping out the step-by-step thought process of students during problem-solving, locating the cause of their error, and providing effective questions without revealing the solution. Manually generating Socratic questions can be a cognitively demanding and time-consuming task for instructors. Several recent works proposed to automatically generate Socratic questions using LLMs: In math education, \cite{shridhar-etal-2022-automatic} shows that generating a sequence of Socratic sub-questions and prompting students to answer helps them solve math word problems more successfully. In computer science education, \cite{10.1145/3626252.3630799,al-hossami-etal-2023-Socratic} releases a dataset on Socratic questions for student code debugging and provides baselines based on LLM prompting and finetuning. In particular, the authors prompt GPT-3.5-turbo and GPT-4 \cite{bubeck2023sparks} in a chain-of-thought manner \cite{wei2022chain} to generate Socratic questions. A human study by the authors shows that the generated questions can sometimes be invalid in several different ways, including being irrelevant to the problem, repetitive of earlier dialogue turns, or too direct and revealing the solution prematurely, which may hamper students' learning processes. Since GPT models are proprietary and expensive, the authors also attempt to fine-tune the open-source Flan-T5 model \cite{chung2022scaling}; however, doing so proves to be ineffective due to its insufficient scale and the pretraining procedure used. 

In this paper, we propose a method to improve the validity of automatically generated Socratic questions using open-source LLMs. Our method is inspired by recent developments in reinforcement learning with AI feedback (RLAIF) \cite{lee2023rlaif}; our method consists of two phases, data augmentation and preference optimization. Specifically, our contributions are as follows: 

\begin{itemize}
    \item To the best of our knowledge, this work is the first to introduce a data augmentation method to create negative samples, i.e., invalid questions, to help us train LLM-based Socratic question generation methods. 
    \item We use the preference information in the dataset, i.e., pairs of valid and invalid Socratic questions, to optimize Llama 2 \cite{touvron2023llama}, an open-source LLM, using direct preference optimization (DPO). \cite{rafailov2023direct}. 
    \item We show that our method using the Llama 2-7B model outperforms existing state-of-the-art methods that rely on larger, proprietary models such as GPT-3.5 and GPT-4 on the Rouge-L metric and are comparable in terms of BERTScore. We also use a series of case studies to illustrate the quality of Socratic questions we generate and that DPO consistently outperforms supervised fine-tuning (SFT). 
\end{itemize}

\section{Related Work}

\subsection{Question Generation in Education}
In education, question-generation systems are used to create learning materials and problem sets for quizzes and exams. \cite{wang-etal-2021-math} introduces a framework for generating math word problems that incorporates a module for checking the consistency of the word problem generated in terms of the underlying equations that it solves. Our idea of checking the consistency of the synthetically generated samples in data augmentation is inspired by theirs. 
\cite{ashok-kumar-etal-2023-improving} proposes a data augmentation and an over-generate and rank method to fine-tune a language model Flan-T5 \cite{chung2022scaling} to generate questions for reading comprehension. %Their experiments show improvements over the previous state-of-the-art results particularly on questions whose answer is not directly present in the passage, hence requiring some reasoning ability. 
Their data augmentation method prompts a larger LLM to augment the dataset with valid questions (positive examples) corresponding to a passage in the reading comprehension and then uses this augmented dataset for standard fine-tuning of a smaller open-source LLM. Unlike their work, our data augmentation method involves prompting a larger LLM to generate invalid questions (negative examples) to create a preference dataset that we use for performing preference optimization on a smaller open-source LLM. In computer science education, recent works show the effectiveness of LLMs like OpenAI Codex and GPT-4 \cite{sarsa2022automatic,kumar2024using} on generating programming exercise questions, code explanations, and test cases. \cite{10.1145/3626252.3630799,al-hossami-etal-2023-Socratic} introduce a Socratic code debugging dataset, to help a student debug their code along with maximizing the students' learning outcomes. Their experiments with prompting models like GPT-3.5-turbo, and GPT-4 show that these models tend to hallucinate and produce invalid questions. To address this issue, our work builds upon theirs to fine-tune language models to align the generated questions towards ground-truth human preferences and discourage the models from generating invalid questions.

\subsection{Reward/ Preference Optimization}
Fine-tuning language models to align with human preferences has proven to be beneficial in various natural language processing tasks \cite{kreutzer-etal-2018-reliability,stiennon2020learning,ziegler2019fine,ouyang2022training}. Traditional methods first learn a reward model using a dataset of human preferences and optimize the language model for the downstream task using the rewards obtained from the reward model with reinforcement learning (RL) algorithms such as PPO \cite{schulman2017proximal}. There are two drawbacks to this method. First, it is hard to obtain a dataset of human preferences as it is an expensive and sometimes cognitively demanding task. To address this issue, RLAIF procures rewards from an AI system, such as an LLM, and has become a scalable and cheaper alternative \cite{lee2023rlaif}. Second, although preference optimization of LLMs using RL algorithms like PPO is effective, it is significantly more challenging and time-consuming than traditional supervised learning as it involves tuning multiple LLMs and sampling rewards in real time. To address this issue, the DPO method \cite{rafailov2023direct} optimizes a language model to a preference dataset in an RL-free manner by formulating the problem as a binary classification task. %DPO significantly reduces the train time and complexity while maintaining similar or even higher performance than traditional PPO methods. 

In the domain of education, \cite{shridhar-etal-2022-automatic} proposes a reward-based method to generate Socratic sub-questions to solve math word problems. Similar to our method they define reward characteristics like fluency, granularity, and answerability to prefer sub-questions that have these desired characteristics. They use REINFORCE \cite{williams1992simple} a popular RL algorithm to optimize their model by sampling rewards from external systems in real time. Our method is different from theirs as we first prompt an LLM to generate invalid Socratic questions (negative examples) to construct a preference dataset. We then use this fixed dataset to tune an open-source LLM in an RL-free method, i.e., using DPO which makes the training more stable and less complex. \cite{hicke2023chata} proposes a DPO-based method for fine-tuning LLama 2 \cite{touvron2023llama} for question-answering on a dataset of Piazza posts for an introductory programming course. They create a proxy preference dataset by using the edit history of Piazza posts by preferring the final versions of answers as opposed to the earlier versions. However, the setting of their work is different from ours as we focus on Socratic question generation and propose a method to create the preference dataset using data augmentation. \cite{scarlatos2024improving} propose a method to perform DPO on LLama 2 for the task of feedback generation to help students solve mathematics word problems. To create preference pairs they prompt LLMs like Codex \cite{chen2021evaluating} and GPT-3.5 turbo to generate bad feedback and rate the feedback based on a pre-defined rubric using GPT-4. Our problem setting is different from theirs as we focus on the programming education domain and for our task the LLM needs to provide a series of step-by-step feedback in the form of a dialogue-based interaction through Socratic questions instead of just providing the feedback once for a given problem.

\section{Problem Definition and Dataset}

We study the problem of Socratic question generation in conversations between a \emph{Student} and an \emph{Instructor}, where the Instructor's goal is to guide the Student through the process of solving a problem. Concretely, our goal is to generate Socratic questions at a particular dialogue turn for the instructor during the conversation, given the dialogue history and contextual information about the problem the Student is trying to solve and their solution.

In this work, we use the dataset for code debugging introduced in ~\cite{10.1145/3626252.3630799,al-hossami-etal-2023-Socratic}. The dataset is based on didactic conversations between a Student and an Instructor, where the Student is a novice programmer tasked with writing a program for a given problem. The dataset consists of the Student's buggy code submissions along with a dialogue between the Instructor and the Student, where the Instructor asks Socratic questions in the form of a conversation to help the Student debug their code. The conversation consists of dialogue turns with each Instructor utterance being a collection of several possible ``ground-truth'' Socratic questions at that dialogue turn. The dataset also contains metadata including the problem statement, the test cases, the bug description, and code fixes to resolve the bug. In total, there are 38 problems with more than 50 different bugs in student solutions, and conversations centered around these buggy codes containing more than 1900 dialogue turns. The dataset is split into two subsets, a train set and a test set which contain 135 and 16 dialogues, respectively, spread across different problems. 

\section{Proposed method}
\begin{figure*}       
    \centering
    \includegraphics[width=0.6\textwidth]{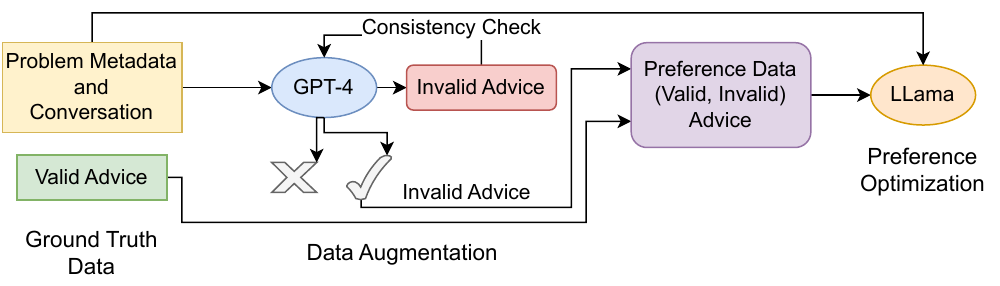}
    \caption{Illustration of our method for LLM-based Socratic question generation, which consists of two phases, data augmentation, and preference optimization.}
    \label{fig:flowchart}
\end{figure*}

In this section, we describe our method for the task of Socratic question generation. Our method involves two phases: First, data augmentation, and second, preference optimization, as shown in Figure \ref{fig:flowchart}. 

\subsection{Data Augmentation}

Inspired by methods in RLAIF \cite{lee2023rlaif}, we augment the dataset with invalid Socratic questions constructed by prompting GPT-4 \cite{bubeck2023sparks}, which provides realistic negative samples for LLM-based question generation methods to train on. We follow the method described in \cite{ashok-kumar-etal-2023-improving} to prompt an LLM to generate synthetic data and employ another instance of the LLM for checking the quality/consistency of the generated synthetic data. Following the definition mentioned in \cite{10.1145/3626252.3630799}, invalid Socratic questions fall into the four following categories: 
\begin{itemize}
    \item \textbf{Irrelevant} questions that are not useful for the student, as they shift focus from the actual bug, which may confuse the student.
    \item \textbf{Repeated} questions that have already been asked in previous dialogue turns, which are meaningless to the student.  
    \item \textbf{Direct} questions that directly reveal the bug to the student, which do not prompt students to think and may hinder their learning process. 
    \item \textbf{Premature} questions which prompt the student to make code fixes before identifying the bug, which may confuse the student. 
\end{itemize}
%Irrelevant utterances shift focus from the actual bug thus confusing the student. Repeated utterances have already been asked or answered in previous dialogue turns. Direct utterances reveal the bug in the code too early thus hampering the learning process of students. Premature utterances prompt the Students for code changes before the actual bug is identified thereby causing confusion in the students. 

To generate invalid questions via an LLM, we construct a few-shot prompt that consists of 1) the definition of the categories as mentioned above and 2) an in-context example for each of the invalid question categories detailed above. Our prompt encourages the model to reason using a chain-of-thought method, by first generating the ``reasoning process/logic'' behind an invalid question, followed by the question \cite{wei2022chain}. We generate invalid questions corresponding to all four categories at every dialogue turn where the ground truth is provided. 

Following \cite{ashok-kumar-etal-2023-improving,wang-etal-2021-math}, we use a consistency checking step where we prompt GPT-4 to check the consistency of the generated questions to filter out inconsistent questions from the augmented dataset. Inconsistent questions are those that do not belong to any of the invalid categories listed above. We pose the consistency checking step as a classification task where GPT-4 predicts a label for each generated question over six categories, including the four invalid categories and two additional categories: ``good'' and ``incorrect''. Good questions are acceptable Socratic questions at that particular dialogue turn and cannot be used as negative samples. Incorrect questions are unrelated to the problem and the dialogue itself and are often erroneous due to LLM hallucination, which provides little value as easy-to-tell negative samples. 
To maintain high data quality of our preference dataset, we discard all samples that are predicted as ``good'' or ``incorrect'', to get the final set of synthetically generated invalid questions. 

Finally, we construct a preference dataset consisting of 2500 tuples of valid and invalid Socratic questions. In the preference pairs, valid questions are taken from the ground truth questions in the original dataset, while the invalid questions are generated synthetically as described above. Each valid question from the original dataset is paired with every synthetically generated invalid question of all categories to form the augmented dataset. 

\subsection{Preference Optimization}
In this step, we fine-tune an open-source LLM, Llama 2 \cite{touvron2023llama} for Socratic question generation using DPO \cite{rafailov2023direct}. The first step is to perform SFT, i.e., we use the original dataset, $D$, as is to fine-tune LLama 2 for Socratic question generation. For a given conversation in the train set, we first split the dialogue into constituent dialogue turns. The input to LLama 2 is a prompt ($p$) that consists of a systems message that instructs the LLM to generate a Socratic question, the problem metadata, and the current dialogue history (between the Student and the Instructor). The output is the valid Socratic question ($q_\text{v}$) corresponding to that dialogue turn in the dataset. In the cases where multiple Socratic questions were given for a dialogue turn, we treat each one as a different output associated with the same input for fine-tuning LLama 2. As shown in Equation \ref{eq:sft}, the simple SFT step learns a reference policy $\pi_\text{ref}$ by minimizing the loss $\mathcal{L}_{SFT}$, which serves as the starting point for preference optimization.

The second step is to perform preference optimization where we fine-tune Llama 2 on the preference dataset, $D_P$, that we obtain from the data augmentation phase, using the same prompt, $p$, as input that was used for SFT, but with two outputs: the valid question $q_\text{v}$ and the invalid question $q_\text{iv}$, for that dialogue turn. As shown in Equation \ref{eq:dpo}, this preference optimization step learns a human preference-aligned policy $\pi_\theta$, given the reference policy $\pi_\text{ref}$ obtained from Equation \ref{eq:sft}, by formulating the task as a binary classification task, minimizing the negative log-likelihood loss $\mathcal{L}_{DPO}$, where $\sigma$ is the Sigmoid function. This minimization leads to learning $\pi_\theta$, by increasing the likelihood of the valid question and decreasing the likelihood of the invalid question while remaining close to the reference policy $\pi_\text{ref}$ which is governed by the hyperparameter $\beta$. Here $\theta$ is the parameters of the preference-aligned policy which is simply the parameters of the neural network, in our case LLama 2.

\begin{align}
\mathcal{L}_{\text{SFT}}(\pi_{\text{ref}}) = - \mathbb{E}_{(q_{\text{v}}, p)\sim D} [\log{\pi_{\text{ref}}(q_{\text{v}} | p)}] \label{eq:sft}
\end{align}

\begin{multline}
\mathcal{L}_{DPO}(\pi_{\theta};\pi_{\text{ref}}) = \\
- \mathbb{E}_{(q_{\text{v}}, q_{\text{iv}}, p)\sim D_P} \bigg[ \log{\sigma(\beta\log{\frac{\pi_{\theta}(q_{\text{v}} | p)}{\pi_{\text{ref}}(q_{\text{v}} | p)}}} \\
- \beta\log{\frac{\pi_{\theta}(q_{\text{iv}} | p)}{\pi_{\text{ref}}(q_{\text{iv}} | p)}}) \bigg] \label{eq:dpo}
\end{multline}

\section{Experimental Settings}

In this section, we detail the implementation setup, methods compared, and metrics used to evaluate our Socratic question generation method.

\noindent\textbf{Implementation details}. 
In the data augmentation phase, we query OpenAI's\footnote{\url{https://openai.com/}} GPT-4 using a rate-based API. We set the temperature of the GPT-4 model to 0.5 to encourage moderate randomness in the outputs. For the consistency checking GPT-4 model, we use a temperate of 0 to maintain determinism. In the preference optimization phase, we use Code-Llama (7B) \cite{roziere2023code} pre-trained for instruction following tasks, particularly on code data\footnote{\url{https://huggingface.co/codellama/CodeLLama-7b-hf}}.
We load our Code-Llama model in an 8-bit configuration and train using QLora \cite{dettmers2023qlora} with the \emph{peft\footnote{\url{https://huggingface.co/docs/peft}}} HuggingFace library to facilitate efficient fine-tuning.  
For the SFT step, we fine-tune the model for 5 epochs with a learning rate of 3e-5, and a batch size of 2 by accumulating gradients for creating a virtual batch size of 64 which takes about 10 hours to train on a single Nvidia A6000 GPU. For the DPO step, we fine-tune the model for 1 epoch with a learning rate of 3e-5 and a $\beta$ (which denotes the KL-loss \cite{joyce2011kullback} between the preference policy learned and the reference SFT policy) of 0.1, with a batch size of 2, which takes about 6 hours to train. For the DPO experiments, we carry out a grid search using hyperparameters learning rate as 1e-5, and 3e-5, $\beta$ of 0.1, and 0.5 and number of epochs as 1 and 2 to arrive at the best-performing hyperparameters as mentioned above. 

\noindent\textbf{Methods}. 
As baselines, we perform zero-shot prompting of the LLama 2 Chat model\footnote{\url{https://huggingface.co/meta-llama/Llama-2-7b-chat-hf}} \cite{touvron2023llama}, denoted by \textbf{LLama}, to generate all possible Socratic questions for the current conversation turn. We also prompt LLama 2 in a chain-of-thought \cite{wei2022chain} manner to first generate the current student misconceptions and then generate the Socratic questions, denoted by \textbf{LLama (CoT)}. 

To decode our trained (SFT and DPO) LLM, we use two decoding techniques, greedy and nucleus sampling, with a $p$ value of 0.9 temperature of 1, and a number of return sequences of 5. We refer to these methods coupled with the trained SFT method as \textbf{SFT Greedy}, \textbf{SFT Sample-5}, and similarly for the DPO methods. Greedy decoding takes 30 minutes to complete, whereas Sample-5 takes an hour.

\noindent\textbf{Metrics}. 
To measure the similarity between the generated Socratic questions and the ground truth questions, we use two commonly used evaluation metrics in natural language generation tasks: \textbf{BERTScore} \cite{bert-score} based on the DeBERTa language model \cite{he2021deberta}, which measures the semantic similarity, and \textbf{Rouge-L} \cite{lin2004rouge}, which measures n-gram overlap based on the longest common subsequence (LCS). In addition, the dataset we use \cite{10.1145/3626252.3630799,al-hossami-etal-2023-Socratic} provides multiple ground truth Socratic questions at each dialogue turn. To measure the similarity between a set of $m$ LLM-generated questions with a set of $n$ ground truth questions, we adopt the process used in \cite{10.1145/3626252.3630799}, which uses Edmond Blossom algorithm \cite{galil1986efficient} to find the maximum matching in a complete bipartite graph between the two sets with a total of $mn$ edges, where the weight of each edge is computed using one of the metrics mentioned above. This step guarantees that every ground-truth question corresponds to, at most, one LLM-generated question, inhibiting semantically equivalent LLM generations from artificially inflating the metric scores. The number of True Positives (TP) is the total sum of the weights of all edges in the optimal matching. False Positives (FP) are calculated by summing the difference between every weight of an edge in the matching with 1. Any unmatched LLM-generated question counts 1 towards False Positive. Similarly, any unmatched ground truth question counts 1 towards False Negative (FN). The TP, FP, and FN values are used to compute the precision, recall, and F1 score for a particular metric. The metric penalizes over-generated LLM questions that do not match with any ground truth questions by classifying them as an FP, thus decreasing the precision.

\section{Results and Discussions}

\begin{table}[t]
\centering
\small
\caption{Performance comparison across different methods. All GPT baseline results are reported in \cite{10.1145/3626252.3630799}. Boldface represents the highest value/s for that column.}
\label{tab:results}
\begin{tabular}{c@{}|ccc|ccc}
\hline
Method & \multicolumn{3}{c|}{Rouge-L} & \multicolumn{3}{c}{BERTScore} \\ \cline{2-7} 
       & P & R & F1          & P & R & F1       \\ \hline
GPT-3.5 & 21.0 & 14.3 & 17.0 & 56.0 & 43.5 & \textbf{48.9} \\
GPT-3.5 (CoT) & 20.3 & 9.7 & 12.0 & 61.7 & 35.8 & 41.6 \\
GPT-4 & 14.1 & 23.3 & 17.6 & 35.4 & 62.6 & 45.2 \\
GPT-4 (CoT) & 5.2 & 26.6 & 8.1 & 12.6 & \textbf{64.8} & 19.5 \\  \hline
LLama & 12.8 & 18.6 & 13.2 & 36.0 & 48.3 & 35.9 \\
LLama (CoT) & 13.7 & 15.5 & 13.2 & 42.3 & 49.0 & 41.0 \\  \hline
SFT Greedy & 29.7 & 13.4 & 17.2 & 61.8 & 29.3 & 36.8 \\ 
DPO Greedy & \textbf{30.6} & 13.3 & 17.1 & \textbf{65.9} & 32.7 & 40.3 \\  \hline
SFT Sample-5 & 14.1 & 26.0 & 17.1 & 32.1 & 62.9 & 41.1 \\
DPO Sample-5 & 15.1 & \textbf{27.9} & \textbf{18.3} & 34.8 & \textbf{64.3} & 42.0 \\ \hline
\end{tabular}
\end{table}

In the consistency checking step of the data augmentation phase, we see that 72\% of the generated questions are considered for the preference dataset creation as 27\% of the generated questions are classified as ``good'' and 1\% as ``incorrect''. This result shows that GPT-4 is more prone to generate ``good'' questions for particular dialogue turns than incorrect questions that do not relate to the problem and the dialogue.

For the task of Socratic question generation, Table~\ref{tab:results} shows the comparison between different methods on the metrics defined for our task. All the GPT-3.5 and GPT-4 results are taken from prior work \cite{10.1145/3626252.3630799}. We observe that GPT-4 (CoT) has the highest recall and yet the lowest F1 score. This observation is because, GPT-4 generates a large number of Socratic questions a few of which are similar to the ground truth questions, however, a significant fraction of the generated questions do not correspond to any ground truth questions, hence being labeled as false positive, thus decreasing the precision. \cite{10.1145/3626252.3630799} also carry out manual analysis to show that GPT (CoT) outputs are the best despite having low F1 scores. This observation can be attributed to the fact that GPT (CoT) has the highest recall among all other GPT methods and hence better corresponds to the ground truth questions.

For the baseline methods that use zero-shot LLama prompting, we observe that LLama (CoT) is the best, which shows that chain-of-thought prompting to first generate the students' current misconceptions followed by the Socratic questions is effective. Among the preference optimization experiments, we see that DPO consistently outperforms SFT. We also observe that the LLama (CoT) performs as well as DPO Greedy in terms of BERTScore F1 as LLama (CoT) generates a higher number of Socratic questions whereas the DPO Greedy method just generates one. Hence, the recall of the DPO Greedy method is lower than that of LLama (CoT). Among decoding variants, we see that the Sample-5 method is better than the Greedy method highlighting the importance of sampling multiple possible Socratic questions instead of just one.

Overall, we see that our preference-optimized models with DPO give the best Rouge-L scores for all precision, recall, and F1 scores with DPO Greedy having the highest precision and DPO Sample-5 having the highest recall and F1 score among all the methods. DPO Greedy has the highest BERTScore precision, whereas DPO Sample-5 has a recall comparable to the best GPT method, GPT-4 (CoT). These results suggest that the DPO-optimized LLama 2-7B model is better than (or as effective as) much larger models like GPT-4 (25 times larger) for Socratic question generation.

\begin{table*}[t]
\centering
\footnotesize
\caption{An example of invalid Socratic questions generated from GPT-4 for a given conversation, which we use to augment the dataset.}
\label{tab:data-aug}
\scalebox{.85}{
\begin{tabular}{p{2.5cm}p{11cm}}
\toprule
Problem & Write a function ``top\_k(lst: List[int], k: int) -> List[int]'' that returns the top k largest elements in the list. You can assume that k is always smaller than the length of the list. \\ 
& Example Case: top\_k([1, 2, 3, 4, 5], 3) => [5, 4, 3]; top\_k([-1, -2, -3, -4, -5], 3) => [-1, -2, -3] 
\\ \midrule
% Buggy Code & def top\_k(lst, k): \\
% & \hspace{0.5cm}result = [] \\
% & \hspace{0.5cm}for i in range(k): \\
% & \hspace{1cm}result.append(max(lst))\\
% & \hspace{1cm}lst.pop(max(lst))\\
% & \hspace{0.5cm}return result \\ \midrule
Bug Description & The function removes the element at index `max(lst)' instead of removing an element equal to `max(lst)'. Consequently, the function throws an IndexError on line 5 when a removed value in `lst' is greater than the length of `lst'. \\ \midrule
Bug Fixes & On line 5, replace `lst.pop(max(lst))' with `lst.remove(max(lst))' \\ \midrule
Conversation & \textbf{Student}: Hi. I am confused. My code doesn't seem to work. Can you help? \\
& \textbf{Instructor}: Hello. Sure, let's see. Do you know what might be the issue? \\
& \textbf{Student}: I think the problem is with the `.pop()' method. It seems to have issues with indexing. \\ \midrule
Ground Truth & 1. Ok, no worries. Let's review your code line by line. Could you please explain it to me? \\
 & 2. Let's start with a simple example. What is the output of the following code snippet: `top\_k([1, 2, 3, 4, 5], 3)'? \\
& 3. Could you please explain what line 5 in your code does? \\
& 4. Let's look into the Python documentation. Can you describe what the `.pop()' method does? \\ \midrule
Invalid Generated & \textbf{Irrelevant}: What happens if you enter an empty list as the input? \\
Questions & \textbf{Repeated}: Do you know what might be the issue? \\
& \textbf{Direct}: Are you sure you should be using the pop() method to remove the maximum element from the list? \\
& \textbf{Premature}: Have you considered using the remove() method instead of pop()? \\ 
\bottomrule
\end{tabular}
}
\end{table*}

\begin{table*}[]
\centering
\small
\caption{Socratic questions generated by different methods.}
\label{tab:questions-methods}
\scalebox{.85}{
\begin{tabular}{p{2cm}p{13cm}}
\toprule
Problem & Write a function `factorial(n:int) -> int' that computes the factorial n! of a natural number n, which is defined mathematically as: $0! = 1$ and $n! = n \times (n - 1)!$. Additionally, if the input integer n is negative the function should return 0. \\ 
& Example Cases: factorial(-1) => 0; factorial(0) => 1; factorial(1) => 1; factorial(5) => 120
\\ \midrule
% Buggy Code & def factorial(n):\\
%     & \hspace{0.5cm}if n < 0: \\
%     & \hspace{1cm}return 0 \\
%     & \hspace{0.5cm}fact = 1 \\
%     & \hspace{0.5cm}for i in range(n): \\
%     & \hspace{1cm}fact = fact * i\\
%     & \hspace{0.5cm}return fact \\ \midrule
Bug Description and Fix & On line 6, `fact' is first multiplied with 0 in the first iteration of the for loop and hence remains 0 throughout the loop. To fix the bug, replace `i' with `(i + 1)' in line 6 and replace `range(n)' with `range(1, n + 1)' in line 5.\\ \midrule
%Bug Fixes & Replace `i' with `(i + 1)' in line 6. Replace `range(n)' with `range(1, n + 1)' in line 5. \\ \midrule

Conversation & \textbf{Student}: Hi! I implemented the factorial function but it doesn’t work and I do not know why. Can you help? \\
& \textbf{Instructor}: Sure. Can you tell me for what values of n it fails and what values it returns in those cases? \\
& \textbf{Student}: For n = 1 or larger it returns the same value, 0. \\ \midrule
Ground Truth & 1. Let's see what happens when `n' is 1. What is the first value that is assigned to variable `i' in line 5? \\
 & 2. Let's see what happens when `n' is 1. Before line 6 is evaluated in the first iteration of the for loop, what are the values of the variables `fact' and `i'? \\
& 3. Let's see what happens when `n' is 1. Can you insert a new line between lines 5 and 6 that prints the values of the variables `fact' and `i'? \\ \midrule
LLama & So, it seems that the issue is that the value of `fact' is not being updated correctly in the loop. Can you tell me why do you think `fact` is not being updated? \\ \midrule
LLama (CoT) & Can you explain why you think the bug is in that specific line? \\ \midrule
SFT Greedy & What is the value of `i' in the first iteration of the for loop? \\ \midrule
DPO Greedy & Okay. Let's start with `n = 1'. What do you think `fact' is on line 6? \\ \midrule
SFT Sample-5 & 1. What is the value of `fact' at line 6 when `i' is 0? \\ 
& 2. Let’s take a closer look at the line where `fact' is multiplied. Could you explain the steps you take to compute that multiplication? \\
& 3. 0 is not a factorial, right? What do you think the first step should be? \\
& 4. What does this tell us about the implementation? \\
& 5. Are you multiplying the initial `fact' value of 1 by `i' at each step? \\ \midrule
DPO Sample-5 & 1. Alright, so for any value of `n' that we pass to the function, what is the output of line 6? \\ 
& 2. Can you tell me what value of `i' is returned in line 6 when `n = 1'? \\
& 3. Okay. Let's look at `factorial(1)' and `factorial(2)'. What are the values of `i' and `fact' when it goes into the for loop in those cases? \\
& 4. Okay, let's go line by line. Let's start from line 5. What do we see? \\
& 5. Okay. Can you tell me the steps for computing `factorial(n)' where `n' equals 1? \\ \bottomrule
\end{tabular}
}
\end{table*}

\begin{table}[]
\centering
\footnotesize
\caption{Buggy codes corresponding to the case studies}
\label{tab:buggycode}
\begin{tabular}{|c|c|}
\hline
\begin{minipage}[t]{0.5\columnwidth}
\textbf{Code 1:}
\begin{verbatim}
def top_k(lst, k): 
 result = [] 
 for i in range(k):
   result.append(max(lst))
  lst.pop(max(lst))
 return result 
\end{verbatim}
\end{minipage} &
\begin{minipage}[t]{0.4\columnwidth}
\textbf{Code 2:}
\begin{verbatim}
def factorial(n):
 if n < 0:
  return 0 
 fact = 1 
 for i in range(n):
  fact = fact * i
 return fact
 
\end{verbatim}
\end{minipage} \\
\hline
\end{tabular}
\end{table}

\section{Case Study}

We now use a case study to illustrate why our method leads to better Socratic question generation. First, we show an example of invalid Socratic questions generated by our data augmentation phase. Second, we compare different methods for Socratic question generation.

Table~\ref{tab:data-aug} shows an example of the augmented data, i.e., invalid questions generated by GPT-4 for an example problem, which asks students to write code to return the largest \texttt{k} elements in a list. The student's code (Table~\ref{tab:buggycode} Code 1) incorrectly removes elements at index \texttt{max(lst)} as opposed to removing elements equal to \texttt{max(lst)}, thereby causing an \texttt{IndexError}. The potential fix to the code is to replace the \texttt{.pop()} function with \texttt{.remove()}. In the conversation, we see that the student knows the problem lies in their use of \texttt{.pop()}. The ground truth Socratic questions for this dialogue turn are highly generic, asking the student to review the code line by line, apply an example test case, or do further reading on Python documentation. We see that the four types of invalid questions generated by GPT-4 are: the \emph{irrelevant} question is out of context and does not help the student understand the bug in their code. The \emph{repeated} question has already been mentioned by the instructor. The \emph{direct} questions reveal the problematic function \texttt{.pop()} and do not give room for the students to discover the problem themselves. The \emph{premature} question directly suggests a code change to replace the \texttt{.pop()} with \texttt{.remove()} function even before the student has realized the actual bug. These diverse examples of invalid questions serve as good training data to let an LLM know what kinds of invalid questions it should avoid generating. 

Table \ref{tab:questions-methods} shows the questions generated by different methods for an example problem that asks students to write code that returns the factorial of a number. The student's code (Table~\ref{tab:buggycode} Code 2) has some indexing errors with the \texttt{range} function, resulting in the \texttt{fact} variable being multiplied by 0 when the loop starts, hence resulting in the output of 0 no matter what the input is. In the conversation, we see that the ground truth questions encourage the student to debug the code by printing the value of lines \texttt{5} and \texttt{6} to examine the variables \texttt{fact} and \texttt{i} along with the role of \texttt{range(n)}. The LLama output is very verbose and directly reveals that the problem is in the updation of the `fact' variable. The LLama (CoT) output is very vague as it does not refer to the exact line of the code. The SFT Greedy output correctly asks the student the value of \texttt{i} but does not provide more details, such as the value of \texttt{n}. The DPO Greedy method is more helpful since it asks the student to check the value of \texttt{fact} specifically for \texttt{n=1}. The first and fifth questions by SFT Sample-5 are invalid and directly ask the value of \texttt{fact} when \texttt{i} is 0, which directly reveals the problem that \texttt{fact} is always 0. The third and fourth outputs are either irrelevant or repeated. The second question, which asks the student to examine the value of \texttt{fact} is valid since it does not directly reveal the bug. In contrast, most of the DPO Sample-5 questions are valid, since they urge the student to examine the value of \texttt{i} and \texttt{fact} on lines \texttt{5} and \texttt{6} with specific values of \texttt{n}, without directly revealing the bug that \texttt{i} is always 0. Through these comparisons, we see that DPO improves Socratic question generation compared to SFT and that DPO Sample-5 is highly capable of generating valid yet diverse questions.

\section{Conclusions and Future Work}
In this work, we propose a method for Socratic question generation in programming problem feedback scenarios. Our method consists of a data augmentation phase to create a preference dataset by synthetically generating invalid questions according to four possible categories. We then use this preference dataset to fine-tune an open-source LLM, LLama 2-7B, using direct preference optimization (DPO). Our results show that the preference-optimized LLama 2-7B model often outperforms existing state-of-the-art prompting methods (on common text similarity metrics) that rely on much larger GPT models (25 times larger), by avoiding invalid questions after training on the augmented dataset. Our method paves the way toward an open-source, accessible, cheaper, privacy-preserving, yet effective alternative to generating Socratic questions which can improve students' learning outcomes without having to rely on proprietary rate-based API-accessed models like GPT-4. There are several avenues for future work. First, we can develop a technique to differentiate types of invalid Socratic questions and not treat them equally while performing preference optimization. This technique would require us to modify the inherent objective function of DPO to incorporate more than one unpreferred question for a single preferred question, which may give us fine-grained control over the LLM generations. Second, we can experiment with open-source LLMs that are larger than 7B to see whether DPO provides more significant gains over SFT on larger models on the Socratic question generation task. Third, we can perform a systematic human evaluation to compare the performance of our proposed method with other baselines. Also, we can focus on designing an automatic metric (based on LLM prompting \cite{liu-etal-2023-g}) other than Rouge and BERTScore which captures the helpfulness of the Socratic questions without heavily relying on assigning higher scores only to questions that have high lexical overlap with the ground-truth questions. Fourth, we can experiment with alternative preference optimization methods, such as KTO \cite{ethayarajh2024kto} which do not need explicit preference data in the form of pairs of valid and invalid questions. Fifth, we can also explore if Socratic question generation helps in improving other tasks in computer science education like test case generation \cite{kumar2024using} by posing the problem as answering several Socratic sub-questions \cite{shridhar-etal-2022-automatic}. Finally, we can also explore how to make Socratic question generation knowledge-aware, i.e., generating different questions for students with different knowledge states, which can be estimated using the open-ended knowledge tracing method for computer science education \cite{liu2022open}.

\section{Limitations}
Our work proposes a method for preference optimizing open-source LLMs like LLama 2 for the task of Socratic question generation for student code debugging. We use only LLama 2 as the base model for carrying out preference optimization, and not other open-source models like Mistral \cite{jiang2023mistral}. Since our main contribution is the data augmentation and preference optimization method, we use only one of the best models open-source models (LLama 2) to show that our method outperforms state-of-the-art models like GPT-4. Future work can also explore the performance of different open-source models using a variety of optimization methods including our data augmentation and preference optimization method for Socratic question generation. Also, we do not formally analyze any biases that exist in the generated augmenting data or the generated Socratic questions. Future work can focus on measuring such biases to make our methods that use these LLMs more inclusive for all students belonging to different demographics. 

\section{Ethics Policy}
Since our invalid questions are generated using an LLM potential linguistic or cultural bias related to the pre-training of the LLM might be reflected. However, we hypothesize that this bias would be minimal as Socratic questions are goal-driven, concise, and framed in the second-person perspective directed toward the student. Our work focuses on open-source LLMs like LLama for Socratic question generation as compared to rate-based API-accessed models like GPT-4 (which is used only once during data augmentation) which implies that our methods are privacy-preserving and there is minimal chance of leakage of students' confidential data. However, training LLMs like LLama on GPUs like A100 for 10 hours results in the emission of CO2 which might not be environmentally friendly. 

\section{Acknowledgments}

We thank Hunter McNichols, Jaewook Lee, Alexander Scarlatos, and Nigel Fernandez for their helpful discussions around this work. The authors are partially supported by the NSF under grant DUE-2215193.


\bibliography{acl_latex}
\clearpage

\section*{\Large Supplementary Material}

\appendix

\section{Prompts}
\label{sec:prompts}

Table \ref{tab:genprompt} shows the prompt used in the data augmentation phase for generating invalid Socratic questions. Table \ref{tab:classprompt} shows the prompt used in the data augmentation phase for consistency checking to classify the generated Socratic questions into six categories. For brevity, we do not include the few-shot examples, which can be found in the code repository.  \newline

\begin{minipage}{\linewidth}
\small
\centering
\captionof{table}{GPT-4 Prompt for Data Augmentation - Generating Invalid Socratic questions}
\label{tab:genprompt}
\begin{tabular}{|p{0.9\textwidth}|}
\hline
\begin{description}
\item[SYSTEM:] 

You are a ``bad'' instructor (Assistant) who generates ``bad'' Socratic questions (based on a predefined category) to help a student debug his code in a conversation with the student. 

\item[Inputs:] 

1. Problem Description

2. Test Cases

3. Student's buggy code

4. Bug Description 

5. Bug Fixes

6. Conversation so far

\item[Bad Categories:] 

1. \textit{Irrelevant}: Questions that shift focus from the actual bug, and ask something not relevant to the current bug.

2. \textit{Repeated}: Questions already asked or answered in the dialogue.

3. \textit{Direct}: Questions that disclose the bug too early, reducing the learning challenge.

4. \textit{Premature}: Questions that guide learners to code changes before they identify the issue, potentially causing confusion.

\end{description} \\
Generate ``bad'' Socratic questions corresponding to each of the four categories and mention the reasoning for the same. \\
\hline
\end{tabular}
\end{minipage}

\clearpage
\begin{minipage}{\linewidth}
\small
\centering
\captionof{table}{GPT-4 Prompt for Data Augmentation - Consistency Checking}
\label{tab:classprompt}
\begin{tabular}{|p{0.9\textwidth}|}
\hline
\begin{description}
\item[SYSTEM:] 

Your task is to output a probability distribution over the labels describing the category of a Socratic question generated by an assistant. 

\item[Inputs:] 

1. Problem Description

2. Test Cases

3. Student's buggy code

4. Bug Description 

5. Bug Fixes

6. Conversation so far

\item[Bad Categories:] 

1. \textit{Irrelevant}: questions that shift focus from the actual bug, and ask something not relevant to the current bug. 

2. \textit{Repeated}: questions already asked or answered in the dialogue. 

3. \textit{Direct}: questions that disclose the bug too early, reducing the learning challenge.

4. \textit{Premature}: questions that guide learners to code changes before they identify the issue, potentially confusing. NOTE: the difference between direct and premature lies in the fact that premature questions specify code changes related to the bug whereas direct questions just reveal the bug directly.

5. \textit{Good}: Questions that are subtle and naturally flow from the conversation without revealing too much about the bug directly/ suggesting code changes prematurely. 

6. \textit{Incorrect}: Questions that are completely out-of-context and are not related to the given problem at all. 

\end{description} \\
The question whose category is to be determined is labeled as `Assistant Socratic Question' in the dialogue. \newline

Output a dictionary containing the labels as the key and the corresponding probability weights as the values. For example, Output: \texttt{{`Irrelevant': 0.6, `Repeated': 0.2, `Direct': 0.1, `Premature': 0.05, `Good': 0.05, `Incorrect':0}} \\
\hline
\end{tabular}
\end{minipage}

\end{document}